\documentclass[times,review,10pt]{elsarticle}

\usepackage{amssymb}
\usepackage{amsmath}

\usepackage{lineno,hyperref}
\usepackage{booktabs}
\usepackage{gensymb}
\usepackage{array}
\usepackage{framed,multirow}
\usepackage{color}
\usepackage{balance}
\usepackage{setspace}
\usepackage{float}
\usepackage{graphicx}
\usepackage{amsmath}
\usepackage{multicol}
\usepackage{bm}
\usepackage{url}
\journal{Pattern Recognition}

\newcommand{\red}[1]{\textcolor[rgb]{0, 0, 0}{{#1}}}
\newcommand{\fix}[1]{\textcolor[rgb]{0, 0, 0}{{#1}}}
\newcommand{\fixx}[1]{\textcolor[rgb]{0, 0, 0}{{#1}}}

\newcommand{\myparagraph}[1]{\vspace{3pt}\noindent{\bf #1}}
\usepackage[top=4.3cm, right=4.8cm, bottom=4.3cm, left=4.8cm]{geometry}
\begin{document}
	
	\begin{frontmatter}
		
		
		
		\title{\fontsize{14}{16.8}\selectfont One-Shot Novel View and Pose Human Image Synthesis via 3D Prior Guided Diffusion Model}
		
		

		\author[1,2]{Shenjian Gong}
		\ead{shenjiangong@njust.edu.cn}
		\author[1]{Kangkan Wang\corref{mycorrespondingauthor}}
		\ead{wangkangkan@njust.edu.cn}
		\author[1]{Jian Yang}
		\ead{csjyang@njust.edu.cn}
		\author[1]{Shanshan Zhang\corref{mycorrespondingauthor}}
		\cortext[mycorrespondingauthor]{Corresponding author}
		\ead{shanshan.zhang@njust.edu.cn}
		
		\address[1]{PCA Lab, Key Lab of Intelligent Perception and Systems for High-Dimensional Information of Ministry of Education, and Jiangsu Key Lab of Image and Video Understanding for Social Security, School of Computer Science and Engineering, Nanjing University of Science and Technology, Nanjing, 210094 China}
		\address[2]{Advanced Laser Technology Laboratory of Anhui Province, Electronic Engineering Institute, National University of Defense Technology, and Jianghuai Advance Technology Center, Hefei 230037, China}
		
		\begin{abstract}
			This paper addresses the challenge of one-shot novel view and pose human image synthesis. The existing methods transfer the reference human image to a target pose using a set of 2D pose keypoints or synthesize human images based on generalizable human NeRF which uses human model priors to extract point-wise features. However, pose transfer based methods can not handle complex human pose using ambiguous 2D pose as the condition, while generalizable human NeRFs may be inaccurate to recover occluded/invisiable human parts without extracted reliable features. To solve these problems, we propose a novel approach for novel view and pose synthesis from a singe human image via conditional denoising diffusion model. Our diffusion model divides the novel view and pose synthesis problem into a sequence of conditional denoising steps. Specifically, to generate humans with complex and arbitrary poses, we introduce 3D human priors, i.e., 3D normal map and color prompt, as geometry and color conditions into the generation process. By transferring the reference human into the target human with a series of diffusion steps, our diffusion model enables high-quality synthesis including the occluded/invisible parts. Further, we propose a self-reconstruction based customized refinement to enhance fine details when tested on novel persons.
			Experimental results on different public datasets demonstrate that our approach significantly outperforms previous methods and also shows better generalization ability across datasets. The code will be made publicly available at \url{https://github.com/Yankeegsj/3DPGDM}.
		\end{abstract}
		
		

		
	\end{frontmatter}
	
		
		\section{Introduction}
		
		Novel view and pose human image synthesis is an important and hot research topic in the fields of VR/AR, and can enable new techniques in downstream tasks, such as single-image 3D human reconstruction and person re-identification. 
		The challenge in this task is that the appearance, texture and shape details of the generated human must match that in the reference image, and the recovered camera view and human pose also need to be consistent with the inputs.
		There are mainly two mainstream methods in this field:
		generalizable human Neural Radiance Fields (NeRFs) \cite{NHP,MPS_NeRF,mononhr,SHERF} and pose transfer \cite{PIDM,PG2,PPATNet} based methods.

		NeRFs \cite{nerf} encodes each 3D coordinate to its color and density, and synthesizes novel-view images through volume rendering \cite{nerf}. Researchers extended NeRFs to generalizable human NeRFs \cite{NHP,MPS_NeRF,mononhr,SHERF} that synthesizes human images in both novel views and poses from a single reference image. The core idea of these NeRFs-based methods is to first extract pixel-aligned features on the reference image at each 3D coordinate, and then predict the geometry and color of the corresponding coordinate from the extracted features. 3D human priors (e.g, SMPL model \cite{smpl}) are always used to associate observation-to-canonical correspondences and facilitate point-wise feature learning across different humans and poses.
		Thus, these methods are good at predicting accurate texture for visible parts on the reference image, yet blurry or details missed synthesis for those invisible parts, where the pixel-level correspondence between the reference and target images cannot be built and thus the extracted features are inaccurate.

		Pose transfer based methods \cite{VUnet,TIP1,TIP2,TIP3,TIP4} aim to transfer the reference person image to a target pose represented by a set of 2D keypoints. In contrast to NeRF-based methods, which perform pixel-level reconstruction, the pose transfer based methods predict the entire humans based on the learned source-to-target transfer trajectories, ensuring high-quality synthesis for both the visible and invisible human body parts. The most recent work, PIDM \cite{PIDM} disintegrates the complex transfer problem into successive denoising steps with the guidance of 2D human pose. However, PIDM can not reliably recover complex human poses since their used 2D pose is insufficient and ambiguous to represent the target view and pose information. In addition, 
		due to overfitting on the limited training data and failure to capture the high-frequency details, 
		human details may not be preserved in the synthesized target images.
		
		In this work, we aim to get the best of both worlds, and propose a 3D guided diffusion model for novel view and pose synthesis from a singe human image.  
		On one hand, to obtain images of high visual quality, we formulate the synthesis problem as a series of conditional denoising steps, which progressively predicts the clean synthesis conditioned on a reference human image and the target view and pose. 
		On the other hand, to enhance the fidelity of synthesized images, we introduce two kinds of 3D guidance: 3D normal map and color prompt predicted from the 3D SMPL model \cite{smpl}, as more accurate and informative conditions.
		Specifically, the 3D normal map can better model human body shapes and better associate correspondences between the target and reference humans; while the color prompt is used to enhance appearance consistency across different views and poses by exploiting texture information from the reference image. 
		As a result, by integrating the above 3D priors, our generator can handle complex human poses even under severe occlusions and obtain high-fidelity synthesis with an arbitrary target view and pose given a single input image. In addition, to address the commonly existed generalization problem in both generalizable NeRFs \cite{SHERF} and pose transfer based methods \cite{PIDM}, 
		we propose a self-reconstruction based customized refinement, where the generator is updated via reconstructing the input reference image for each novel person.
		In this way, we further improve the synthesis accuracy and enhance the high-frequency details, so that the generated human image is more consistent with the input reference image.

		To summarize, the main contributions of our work are as follows:
		\begin{itemize}
			
			
			
			\item[$\bullet$] 
			Aiming for high-quality one-shot human image synthesis under novel views and poses, we propose to equip a diffusion based generation model with useful 3D guidance: the 3D normal maps are used to precisely model the human body pose and shape so that the correspondences between the reference and target human bodies are better established; the color prompt is used to preserve texture information from the reference image via explicit pixel alignment.
			
			\item[$\bullet$] 
			To better recover details for an unseen novel person, we propose a customized refinement, which updates the generator via reconstructing the input reference image.
			\item[$\bullet$] Our proposed method achieves state-of-the-art performance for one-shot human image synthesis task on two public datasets (i.e., RenderPeople \cite{renderpeople} and THuman \cite{thuman} datasets).
			
		\end{itemize}

		\section{Related Work}
		In this section, we briefly review the recent works on human NeRFs based methods and pose transfer based methods.
		\subsection{Human-specific and generalizable human NeRFs methods}
		Aiming to synthesize both novel-view and novel-pose human image, the prior NeRF-based methods learn a canonical NeRFs in a pose-aligned space (named canonical space), and fetch density and color for the observation-space points by transforming them to the canonical space via a deformation fields.  HumanNeRF \cite{humannerf} computes the deformation field by using the linear blend skinning (LBS) method \cite{smpl}. Ani-NeRF \cite{aninerf} refines a blend weight fields in the skeleton-driven deformation to associate more accurate correspondences between the observed and canonical spaces.
		NeuralBody \cite{Neuralbody} learns the latent codes anchored to a deformable mesh and obtains novel view and pose synthesis by deforming the mesh to observation space. Recent works \cite{humangs1,humangs2,humangs3,humangs4} attempt to improve the quality and speed of human rendering by introducing the newly proposed 3D Gaussian splatting \cite{3dgs,SVAD,Disco4d,SinGS}. Although the synthesis results are promising, these human-specific NeRF methods require re-optimization for each person and can not work on unseen persons. 
		
		Generalizable human NeRFs are proposed to predict NeRFs from point-wise features extracted from the input images. NHP \cite{NHP} introduces a temporal transformer and a multi-view transformer to capture the visual features on the temporal body motion and the multi-view pixel-aligned features at each time, respectively. MPS-NeRF \cite{MPS_NeRF} generates pixel-aligned geometry and color features by fusing extracted image features with two self-attention blocks.
		MonoNHR \cite{mononhr} and SHERF \cite{SHERF} learn generalizable Human NeRFs based on a single image. MonoNHR \cite{mononhr} only explores novel view synthesis, while SHERF \cite{SHERF} combines global, point-level and pixel-aligned features to facilitate informative encoding of the single image. The generalizable NeRFs based methods cannot extract reliable features in the occluded or invisible human parts on the single image, leading to the synthesis of these parts blurry and lacking of details. In this work, we aim to obtain high-quality novel view and pose synthesis for the entire humans including occluded or invisible parts.
		

		\subsection{Pose transfer based methods}
		The pose transfer based methods \cite{pt1,pt2,pt3,pt4,pt5,PPATNet,controllable,ptdiff} aim to synthesize a person’s image with a target 2D pose and a desired appearance consistent with the given reference image. This task has received increasing attention in recent years with the success of conditional GAN \cite{CGAN}.
		PG2 \cite{PG2} concatenates the reference image and pose, and the target pose as inputs and performs pose transfer in a coarse-to-fine manner, which may results in pose misalignment at pixel level. 
		To improve performance for images with large spatial variation, VUNet \cite{VUnet} introduces conditional VAE to disentangle the appearance and the pose conditions.
		PPATNet \cite{PPATNet} predicts the images progressively via a sequence of pose-attentional transfer blocks, making it easier than one-step pose transfer. \fix{A Liquid Warping GAN \cite{liquid} employs a 3D body mesh recovery module to disentangle the pose and shape, and dynamically aggregates source features through spatial attention maps, enabling motion imitation and appearance transfer.}
		The above GAN-based methods are mainly proposed for person image generation with simple poses, and it is difficult for them to model the transformation among complex poses with the elaborately designed architecture.

		Recently, diffusion model \cite{diffusion} has obtained better performance than GAN on the task of conditional image synthesis \cite{dhariwal2021diffusion, GLIDE, CFDG}. 
		PIDM \cite{PIDM} is the first work to successfully divide the complicated pose transfer task into a sequence of simple denoising problems. PIDM \cite{PIDM} generates the target human image conditioned on 2D human pose and obtains good results in easy standing poses. \red{Based on 2D human pose, DisCo \cite{disco} introduces disentangled control to improve the compositionality of dance synthesis and proposes a human attribute pre-training for generalizability to unseen humans.}
		\red{However, the 2D human poses cannot effectively represent the 3D human body shape, and may provide ambiguous view and pose information due to serious occlusions under a single image, leading to dramatically degraded performance especially under complex human poses. Therefore, MagicAnimate \cite{magicanimate} utilizes the guiding condition of a dense pose map instead of a 2D pose skeleton map. 
			\fix{Xue et al. \cite{xuetowards} proposed a multi-condition guidance framework that uses optical flow, depth ordering, and reference pose maps to enhance implicit decoupling of background and character features.}
			Additionally, Human4DiT \cite{Human4DiT} and Champ \cite{champ} introduce 3D normal maps to precisely capture intricate human geometry details. Besides normal maps, Champ also leverages depth maps, semantic maps, and 2D skeleton maps. Nevertheless, acquiring these maps is a time-consuming process, and ensuring the accuracy  and consistency of different conditions is challenging. Consequently, this can potentially compromise the quality of the generated results. 
			In Human4DiT \cite{Human4DiT}, it is stated that their method implicitly associates moving humans of different views and poses through the attention mechanism. The absence of an explicit representation like our proposed color prompts, however, may give rise to obvious artifacts when rendering free-view images. 
		} 
		
		\red{In this work, we employ the conditional diffusion model for one-shot novel view and pose human image synthesis. Moreover, we focus on more efficient utilization of 3D normal maps in the context of synthesizing novel view and pose human images, and introduce both implicit and explicit 3D prior guidance as conditions to improve the synthesis quality.}

		\begin{figure*}[t]
			\begin{center}
				\includegraphics[width=0.95\textwidth]{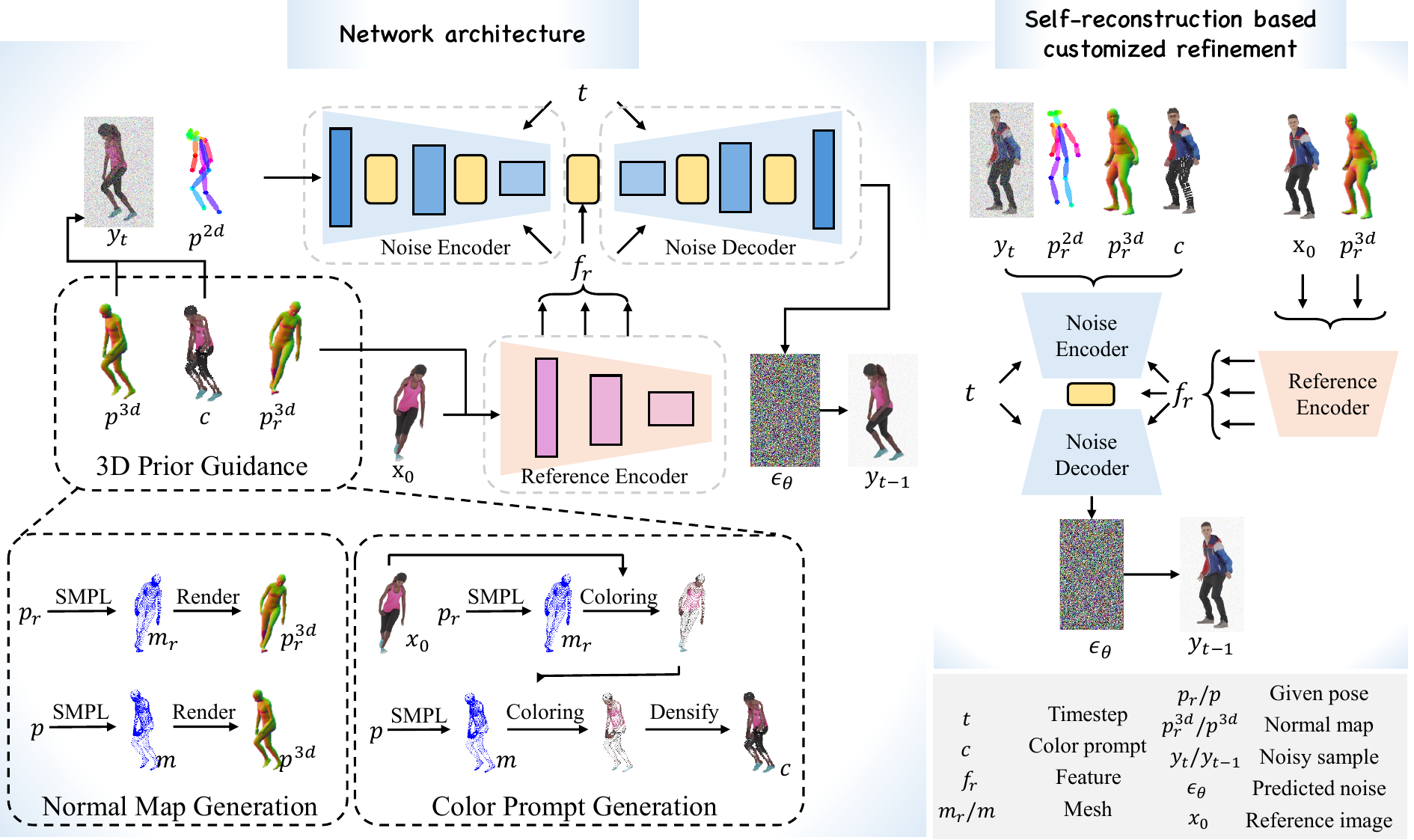}
			\end{center}
			\caption{Overview of our proposed method.
				\textcolor{black}{Given a reference image $x_0$ and the target pose $p^{2d}$, our model predicts the target image $y_0$ via a denoising diffusion model (Sec. \ref{sec:diffusion}). We introduce 3D prior guidance (Sec. \ref{sec:3d}) as additional conditions and a self-reconstruction based customized refinement (Sec. \ref{sec:self-generation}), to enhance image fidelity.}} 
			\label{fig:method}
		\end{figure*}
		
		\section{Proposed Method}
		
		
		
		Fig. \ref{fig:method} shows the overview of our proposed method. The generation model predicts the target image with a novel view and pose via a denoising diffusion model (Sec. \ref{sec:diffusion}), given the reference image along with the target 2D pose as inputs, and our proposed 3D prior guidance (Sec. \ref{sec:3d}) as additional conditions. We further describe our proposed self-reconstruction based customized refinement in Sec. \ref{sec:self-generation}, which is used to enhance recovery details for a novel person.
		
		\subsection{3D human prior guided denoising diffusion model}
		\label{sec:diffusion}
		Given a reference image $\bm{x}_0 \in \mathbb{R}^{3 \times H \times W}$ of a person associated with its camera parameter $\bm{v}_r$ and human pose $\bm{p}_r$ in the form of SMPL parameters, our goal is to synthesize the image $\bm{y}_0 \in \mathbb{R}^{3 \times H \times W}$ of the person in a target pose $\bm{p}$ from target view $\bm{v}$.

		Our human image synthesis framework is based on the conditional denoising diffusion probabilistic model (DDPM) \cite{diffusion}, which consists of $T$ conditional denoising diffusion steps. In the forward diffusion process, noise is gradually added to the sampled target image $\bm{y}_0 \sim q(\bm{y}_0)$, and the backward denoising process aims to learn the reverse mapping. The trained denoising model can convert an isotropic Gaussian noise $\bm{y}_t \sim \mathcal N(0,\bm{I})$ into the target human image with $T$ denoising steps. The forward diffusion is a Markov chain with the following conditional distribution:
		\begin{equation}
			q(\bm{y}_t|\bm{y}_{t-1})=\mathcal N(\bm{y}_t; \sqrt{1-\beta_t}\bm{y}_{t-1},\beta_t\bm{I}),
			\label{equation:add_noise}
		\end{equation}
		where $t\sim [1,T]$, $\beta_t\sim (0,1)$. Let $\alpha_t=1-\beta_t$ and $\eta_t = {\textstyle \prod_{i=1}^{t} \alpha_i}$. We can add the noise $\bm{\epsilon} \sim \mathcal N(0,\bm{I})$ to $\bm{y}_0$ and obtain the noisy sample directly from $q(\bm{y}_t|\bm{y}_0)$ at the $t$-th timestep as:
		\begin{equation}
			\bm{y}_t=\sqrt{\eta_t}\bm{y}_0+\sqrt{1-\eta_t}\bm{\epsilon}.
		\end{equation}
		
		In the reverse denoising pass, we predict the noise via a deep neural network to denoise $\bm{y}_t$ based on our elaborately designed conditions (described in Sec. \ref{sec:3d}). Following PIDM \cite{PIDM}, our adopted network consists of three learnable parts: denoising network, reference encoder and cross-attention module. The denoising network is a UNet-based design composed of a noise encoder and decoder which maps the noisy sample to added noise, and the reference encoder extracts multi-scale features $\bm{f}_r$ from the source image. In our denoising model, the noise encoder concatenates $\bm{y}_t$ with 2D pose $\bm{p}^{2d}$, 3D normal map $\bm{p}^{3d}$ and color prompt $\bm{c}$ as the input, while the reference encoder is fed with $\bm{x}_0$ and $\bm{p}_r^{3d}$. To make the denoising aware of the reference human, we perform layer-wise cross-attention between noise features $\bm{f}^l_n$ from noise encoder/decoder and multi-scale features $\bm{f}^l_r$ in corresponding layers. The layer-wise cross-attention is formulated as:
		\begin{equation}
			\begin{split}
				\bm{Q}&=\bm{\theta}_q^l(\bm{f}_n^l),  \bm{K}=\bm{\theta}_k^l(\bm{f}_r^l), \bm{V}=\bm{\theta}_v^l(\bm{f}_r^l),\\
				\bm{f}_o^l&=\bm{W}^lsoftmax(\frac{\bm{QK}^T}{\sqrt{d}})\bm{V}+\bm{f}_n^l,
			\end{split}
			\label{equation:QKV}
		\end{equation}
		where $\bm{\theta}_q^l, \bm{\theta}_k^l$ and $\bm{\theta}_v^l$ are convolution operators with kernel size of $1\times 1$ applied to corresponding features of $l$-th layer, and $\bm{W}^l$ denotes the learnable cross-attention weights for output features $\bm{f}_o^l$. \fix{In this way, the generation model is able to implicitly build pixel-level human correspondences with the aid of $\bm{p}_r^{3d}$ and $\bm{p}^{3d}$ via performing cross-attention between $\bm{f}_r$ and $\bm{f}_n$. Additionally, the color prompt $\bm{c}$, as a color image of the target human obtained from explicit pixel alignment using the 3D human meshes, provides useful appearance information for synthesizing the target human image. }
		
		For optimization, we define the denoising loss $\boldsymbol{\mathcal{L}}_{D}$ as a standard MSE loss between the added noise and predict one $\bm{\epsilon}_\theta(\bm{y}_t,t,\bm{x}_0,\bm{p}^{3d}_r,\bm{p}^{2d},\bm{p}^{3d},\bm{c})$: 
		\begin{equation}
			\boldsymbol{\mathcal{L}}_{D}= \left \| \bm{\epsilon} -\bm{\epsilon}_\theta(\bm{y}_t,t,\bm{x}_0,\bm{p}^{3d}_r,\bm{p}^{2d},\bm{p}^{3d},\bm{c}))  \right \| ^2.
			\label{equation:denosing loss}
		\end{equation}
		
		Additionally, following previous works \cite{PIDM,vib}, we approximate the posterior $\bm{q}(\bm{y}_{t-1}|\bm{y}_t)$ via the denoising network:
		\begin{equation}
			\bm{p}_\theta(\bm{y}_{t-1}|\bm{y}_{t})=(\bm{y}_{t-1};\bm{\mu}_\theta, \bm{\nu}_\theta),
		\end{equation}
		where the mean $\bm{\mu}_\theta$ is derived from the predicted noise $\bm{\epsilon}_\theta$ and the variance $\bm{\nu}_\theta$ is learned in the training (for details please refer to \cite{vib}). We introduce variational lower-bound loss $\mathcal{L}_{vlb}$ for effective diffusion model learning, defined as:
		\begin{equation}
			\boldsymbol{\mathcal{L}}_{vlb} = 
			\begin{cases} 
				-log(\bm{p}_\theta(\bm{y}_{t-1}|\bm{y}_{t})), & \text{if t=0},\\
				D_{KL}(\bm{q}(\bm{y_t}|\bm{y}_{t-1})||\bm{p}_\theta(\bm{y}_{t-1}|\bm{y}_{t})), & \text{otherwise}.
			\end{cases}
		\end{equation}
		
		Then, the overall optimization objective is defined as follows:
		\begin{equation}
			\mathcal{L}=\mathcal{L}_D+\mathcal{L}_{vlb}.
			\label{equation:loss}
		\end{equation}
		
		Similar to PIDM \cite{PIDM}, our model divides the complex pose transfer problem into a sequence of conditional denoising steps. At the inference stage, we first sample the noise $\bm{y}_T \sim \mathcal N(0, I)$, and progressively sample from $\bm{p}_\theta(\bm{y}_{t-1}|\bm{y}_{t})$ with $T$ denoising steps from t=$T$ to t=1, and finally obtain the synthesized image.

		\subsection{3D human prior guidance}
		\label{sec:3d}
		Although pose transfer based approaches \cite{PIDM,PG2,PPATNet,VUnet} can be applied to novel view and pose person image synthesis, the performance is poor because  
		the guidance provided by $\bm{x}_{0}$ and the target 2D pose $\bm{p}^{2d}$ is rather limited.
		To better guide the generation, we explore to utilize 3D human priors produced from a 3D SMPL model, i.e., 3D normal map and color prompt. Compared with 2D pose, the 3D normal map can better model the human pose and shape and associate human correspondences between the target and reference. The color prompt allows the generator to exploit appearance information explicitly aligned with the target view and pose. In the following, we provide a further description of how to generate 3D human prior guidance used in our method.
		
		
		\begin{figure}[t]
			\begin{center}
				\includegraphics[width=0.95\textwidth]{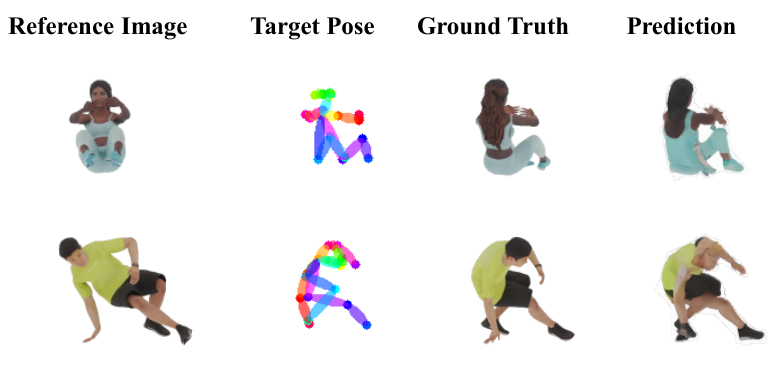}
			\end{center}
			\caption{\textcolor{black}{Failure cases of PIDM \cite{PIDM} under complex target poses with self occlusions.}} 
			\label{fig:2d pose error}
		\end{figure}
		
		\myparagraph{3D normal map}: 
		In the human image synthesis, we expect to handle complex human poses even under severe occlusions. However, due to the weak representation ability of 2D pose, pose transfer based methods have difficulty transferring very complex poses, failing to synthesize high-quality images. 
		As shown in Fig. \ref{fig:2d pose error}, for complex human poses with severe self occlusions, the 2D pose fails to represent the geometry of the whole human body due to lack of depth information, and thus results in inaccurate predictions.
		To provide more informative pose guidance, we introduce 3D mesh geometry as an extra condition in the form of normal map. Given the SMPL parameters of a 3D human $\bm{p}(\bm{\alpha},\bm{\beta})$ consisting of the body pose parameter $\bm{\alpha}$ and the shape parameter $\bm{\beta}$, SMPL model creates a mesh $\bm{M}\in \mathbb{R}^{6890 \times 3}$ with 6890 vertices. As shown in Fig. 
		\ref{fig:method}, we compute the normal map of the 3D human mesh $\bm{M}$ and render it from the input camera view. As the input into the reference encoder, we concatenate $\bm{x}_{0}$ with the normal map of the reference human $\bm{p}_r^{3d}$. In this way, the generation model is able to implicitly build pixel-level human correspondences with the aid of $\bm{p}_r^{3d}$ and $\bm{p}^{3d}$ via performing cross-attention between $\bm{f}_r$ and $\bm{f}_n$.
		
		\begin{figure*}[h]
			\begin{center}
				\includegraphics[width=0.99\textwidth]{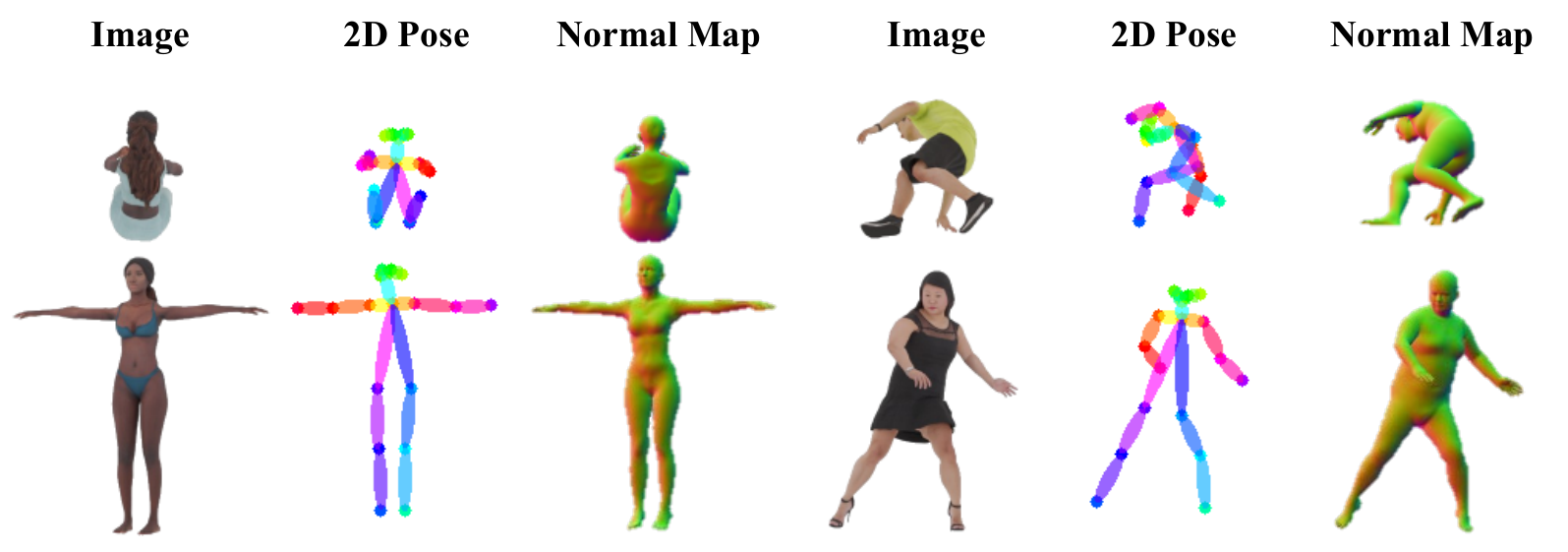}
				\caption{\textcolor{black}{Comparisons of 3D normal map and 2D pose. The 3D normal map is able to represent complex postures more accurate and better represent the human body shape than 2D pose. }}
				\label{fig:2d vs 3d}
			\end{center}
		\end{figure*}
		
		To better demonstrate the advantages of using the 3D normal map, we show some examples in Fig. \ref{fig:2d vs 3d}. 
		In row 1, the two persons are of complex postures with severe self occlusions, where one can hardly deduce the human body shape from the ambiguous 2D pose; in contrast, the 3D normal map with geometric information better represents the human body shape.
		In row 2, the left person is obviously slimmer than the right one, which is not represented by the 2D pose but well reflected by the 3D normal map.
		Therefore, we conclude that the 3D normal map provides much richer information regarding human body shape and thus enables the generator to synthesize humans in a more precise way.

		\begin{figure*}[t]
			\begin{center}
				\includegraphics[width=0.95\textwidth]{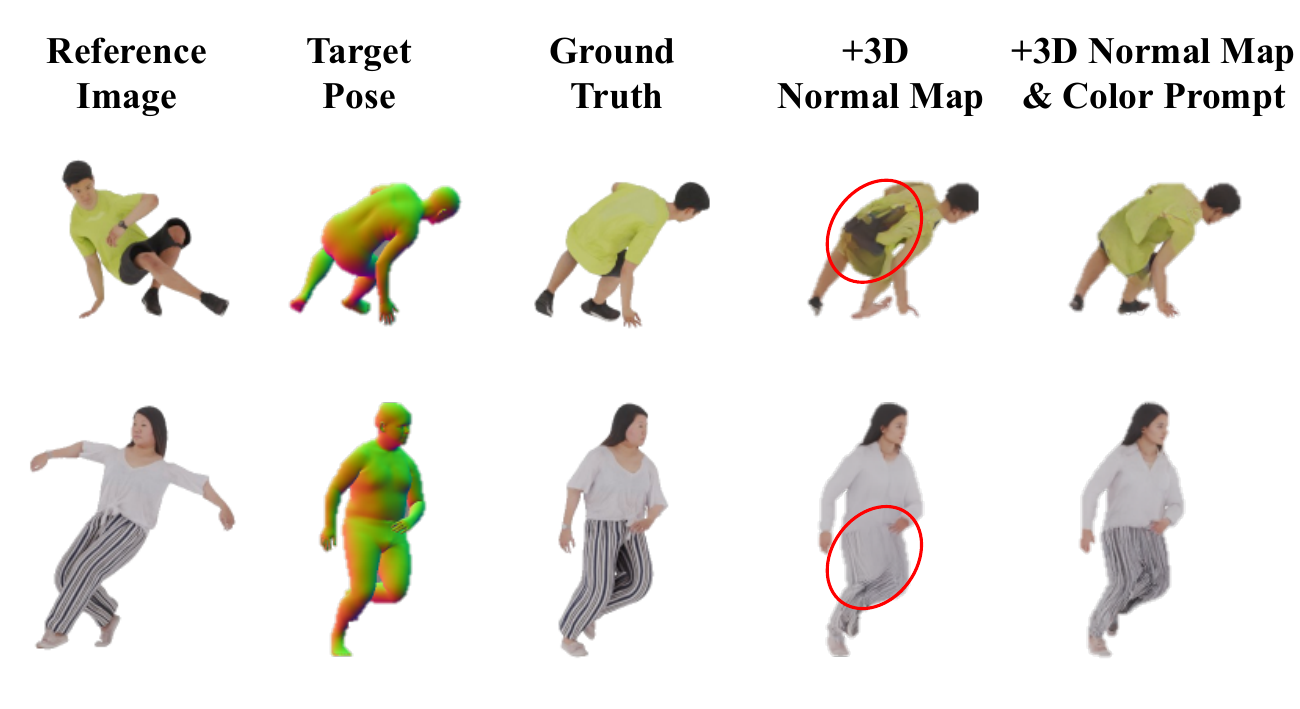}
			\end{center}
			\caption{\fixx{The motivation and effects of color prompt. The 3D normal map ensures geometric consistency, while the color prompt further preserves appearance and texture details. Zoom in for better visualization.}}
			\label{fig:color prompt advantage}
		\end{figure*}
		
		\myparagraph{Color prompt}: 
		As shown in Fig. \ref{fig:color prompt advantage}, the use of a 3D normal map can only ensure that the prediction meets the geometric condition, yet it falls short in generating the desired appearance and texture. 
		Therefore, in addition to the 3D normal map, we also introduce color priors in the denoising process to preserve texture information from the reference image. Examples of this are shown in Fig. \ref{fig:color prompt advantage}.
		To be specific, the generation process of the color prompt $\bm{c}$ is shown in Fig. \ref{fig:method}. To be specific, with the reference human mesh $M_r$ generated from the reference SMPL parameters $\bm{p}_r$, we project all vertices of $\bm{M}_r$ to $\bm{x}_{0}$, so as to obtain the image coordinates on $\bm{x}_{0}$ and take the pixel colors for these projected vertices. Then, the vertex coordinates of $\bm{M}_r$ are replaced with the human mesh $M$ generated from the target SMPL parameters $\bm{p}$, and the corresponding colors are projected to the target camera, exporting a color prompt image $\bm{c}$ in the target view and pose. It is worth noting that we take the visibility of the human body into account when constructing color prompt, and only the visible vertices in the target view are activated as effective points. As the effective points are sparse, we extend the hint color of each point to the neighbouring pixels within a radius of 1 pixel to enhance the density. Actually, $\bm{c}$ is a color image of the target human obtained from explicit pixel alignment using the 3D human meshes, which provides useful appearance information for synthesizing the target human image. 
		
		\subsection{Self-reconstruction based customized refinement}
		\label{sec:self-generation}

		\begin{figure*}[t]
			\begin{center}
				\includegraphics[width=0.95\textwidth]{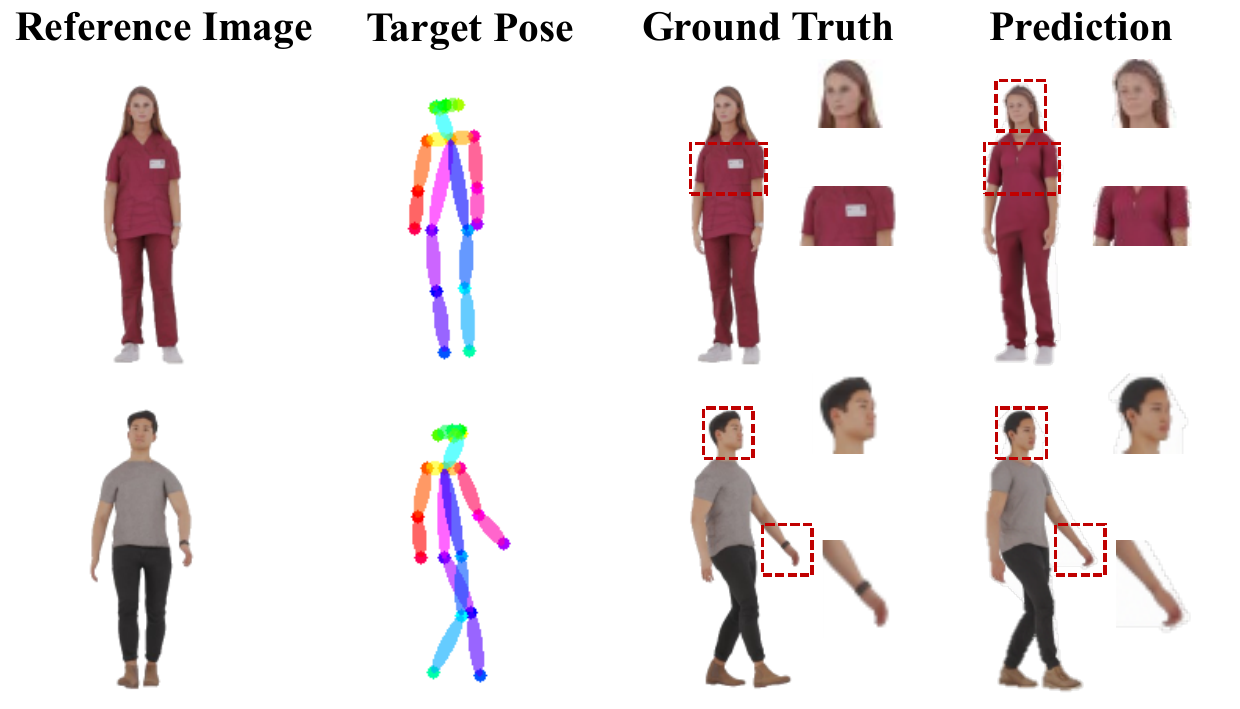}
			\end{center}
			\caption{\textcolor{black}{Failure cases of PIDM \cite{PIDM} with wrong or missing details. Zoom in for better view.}} 
			\label{fig:detail miss}
		\end{figure*}
		
		In the previous subsections, we guide the generation of the denoising model by elaborately designing the geometry and color conditions. However, human details (e.g., facial details or local cloth texture) might not be recovered when generalizing the model to a novel person. Some cases are shown in Fig. \ref{fig:detail miss}. The appearance and color of the generated clothes globally match the reference image. Whereas, in row 1, the predicted face does not match the reference person and the white label on the shirt is missing. Similarly, in row 2, the generated man has a realistic but inaccurate face and loses his watch on his hand. The problem of failing to recover these local or high-frequency details is mainly caused by overfitting on the limited training data.
		
		In order to boost the pixel-level synthesis accuracy and enhance the high-frequency details on the synthesized human that match with the input reference human, 
		we propose a self-reconstruction based customized refinement to better adapt our generator to a novel person at test time.
		To be specific, 
		as shown in Fig. \ref{fig:method}, given pose parameters $\bm{p}_r$ and reference image $\bm{x}_{0}$,  the 3D normal map $\bm{p}_r^{3d}$ is rendered from the reference human mesh $\bm{M_r}$ and the color prompt $\bm{c}$ is generated via coloring $\bm{M_r}$.
		We feed the 3D normal map $\bm{p}_r^{3d}$ and $\bm{x}_{0}$ to the reference encoder. The noise encoder takes $\bm{p}_r^{2d}$, $\bm{p}_r^{3d}$, $\bm{c}$ and noisy sample $\bm{x}_t$ of $\bm{x}_{0}$ at a random time step $t$ as the input. The denoising loss $\boldsymbol{\mathcal{L}}_{D}$ in Eqn. \ref{equation:loss} is then updated as $\boldsymbol{\mathcal{L}}_{D}'$: 
		\begin{equation}
			\boldsymbol{\mathcal{L}}_{D}'= \left \| \bm{\epsilon} -\bm{\epsilon}_\theta(\bm{x}_t,t,\bm{x}_0,\bm{p}^{3d}_r,\bm{p}^{2d}_r,\bm{c}))  \right \| ^2.
			\label{equation:selfgeneration denosing loss}
		\end{equation}
		
		In the refinement, we finetune the generator for novel person to preserve the fine details of the reference human for $\lambda$ steps. In essence, self-reconstruction treats the reference image as target image simultaneously and captures the personalized details during the refinement. Thus, the optimized model can be seen as a customized generator for the novel person. 
		
		\fixx{It is notable that the term ``one-shot" in the title strictly refers to the data requirement in the inference—only one reference sample is provided per target identity. Our self-reconstruction-based refinement only utilizes the input single sample, which is analogous to the test-time adaptation of previous works that employ test-time techniques with few/one-shot data \cite{ranjan2021learning,yang2023one,wu2024test}.}\fixx{By default, our refinement runs 100 steps and takes approximately 90 seconds. And this stage only optimizes the model using the given single reference image without any extra data.}

		\section{Experiments}
		In this section, we first describe the experimental setup, including the datasets, implementation details and evaluation metrics; then we compare our method to the state-of-the-arts; finally, we perform ablation studies of our method.
		\subsection{Experimental setup}
		\myparagraph{Datasets:} Two large-scale human datasets are chosen for evaluation. (1) For RenderPeople \cite{renderpeople}, 
		the rendered images are generated by Hu et al. \cite{SHERF}. For each subject, there are 36 different views. Following SHERF \cite{SHERF}, we sample 450 subjects as training set and 30 subjects for testing. (2) For THuman \cite{thuman}, we render the images with 10 more views than RenderPeople for each subject. We take 191 subjects as training set and 10 subjects for testing. 
		
		\myparagraph{Evaluation metrics:}
		Following previous works \cite{SHERF,MPS_NeRF,Neuralbody}, we mainly adopt $Peak$ $Signal$-$to$-$noise$ $Ratio$  (PSNR) \cite{psnr}, $Structure$ $Similarity$ $Index$ $Measure$ (SSIM) \cite{ssim} and $Learned$ $Perceptual$ $Image$ $Patch$ $Similarity$ (LPIPS) \cite{lpips} as evaluation metrics. PSNR quantifies synthesis accuracy, SSIM calculates the structure similarity, and LPIPS measures perceptual image patch similarity. Following the previous Human NeRF based methods \cite{SHERF,MPS_NeRF,Neuralbody}, we only calculate these metrics in the area of 2D projected human bounding box between the synthesized image and the ground-truth image. Additionally, we take $Fr\acute{e}chet$ $Inception$ $Distance$ (FID) to quantify the realism of the generated images.

		\myparagraph{Implementation details:} For our denoising network, following PIDM \cite{PIDM}, we perform cross-attention in the UNet-like encoder-decoder architecture at some specific resolutions ($32 \times 32$, $16 \times 16$ and $8 \times 8$). The total noising steps of our denoising model are T = 1000 with a linear noise schedule. During training, the timestep $t$ is sampled in [1,1000]. For all experiments, we train our model using $256\times 256$ images and set the background to white color. We train the network with Adam optimizer, and the learning rate is set to $2e^{-5}$. All experiments are conducted on 8 NVIDIA RTX 3090TI GPU with 24GB of VRAM and our codes are implemented with Pytorch. The batch size on each GPU is 4. 
		
		\fixx{For inference, the diffusion process is sampled for 50 steps and takes approximately 3 seconds to generate a $256\times 256$ human image on a single NVIDIA RTX 3090TI GPU. The self-reconstruction based customized refinement is optimized for $\lambda=$100 steps on a single NVIDIA RTX 3090TI GPU, with a runtime of about 90 seconds (executed once per identity rather than per image). Additionally, we recommend enabling the refinement stage for all users to improve the quality of generated images. However, this fine-tuning step can be skipped if users have limited hardware resources or tight time constraints since our method without the refinement step still can yield satisfied results.}
		
		\subsection{Comparisons to the state-of-the-art methods}
		\red{First of all, we compare our method with SoTA generalizable human NeRF based methods (MPS-NeRF \cite{MPS_NeRF} and SHERF \cite{SHERF}) and the top performing pose transfer method (PIDM \cite{PIDM}, DisCo \cite{disco}, PoCoLD \cite{pocold} and Champ \cite{champ}) on the RenderPeople and THuman datasets. It's worth noting that the training data and evaluation strategies employed in pose transfer differ significantly from those used in novel view and pose human image synthesis. To ensure a fair comparison, we train and test these methods in accordance with the experimental setup in the novel view and pose human image synthesis task. 
		}
		
		\setlength{\tabcolsep}{2pt}
		\begin{table}[t]
			
			\renewcommand*{\arraystretch}{1.15}
			\centering
			\begin{tabular}{l|cccc|cccc}
				\toprule
				
				\multirow{2}{*}{Method} &\multicolumn{4}{c|}{Novel View}&\multicolumn{4}{c}{Novel Pose}\\
				\cline{2-9} 
				~ &FID$\downarrow$ &PSNR$\uparrow$ &SSIM$\uparrow$& LPIPS$\downarrow$ &FID$\downarrow$&PSNR$\uparrow$ &SSIM$\uparrow$& LPIPS$\downarrow$ \\
				\hline
				MPS-NeRF &45.12 &20.36 &0.84& 0.16 &47.80& 19.61 & 0.82 & 0.18\\
				SHERF&36.54& 21.37 & 0.88 & 0.12 &38.98& 20.65 & 0.86 & 0.13 \\ 
				\red{DisCo} &\red{21.44}& \red{19.57} & \red{0.81} & \red{0.12}&\red{23.90}& \red{18.89} & \red{0.79} & \red{0.13} \\ 
				PoCoLD &22.16& 18.31 & 0.81 & 0.15 &25.23& 17.60 & 0.79 & 0.16 \\ 
				PIDM &17.96& 19.82 & 0.85 & 0.11&\textbf{19.46} & 19.12 & 0.83 & 0.12 \\ 
				\red{Champ} &\red{20.93}& \red{20.53} & \red{0.82} & \red{0.10}&\red{22.84}& \red{19.21} & \red{0.79} & \red{0.12} \\
				\hline
				Ours &\textbf{17.05}& \textbf{23.29} & \textbf{0.91} & \textbf{0.07} &19.53 & \textbf{22.36} & \textbf{0.90} &\textbf{0.07}\\ 
				\toprule
			\end{tabular}
			\caption{\textcolor{black}{Comparison to the state-of-the-art methods on the RenderPeople dataset. \fixx{$\uparrow$: higher is better. $\downarrow$: lower is better.}}}
			\label{table:STOA}
		\end{table}

		\begin{figure}[t]
			\begin{center}
				\includegraphics[width=0.95\textwidth]{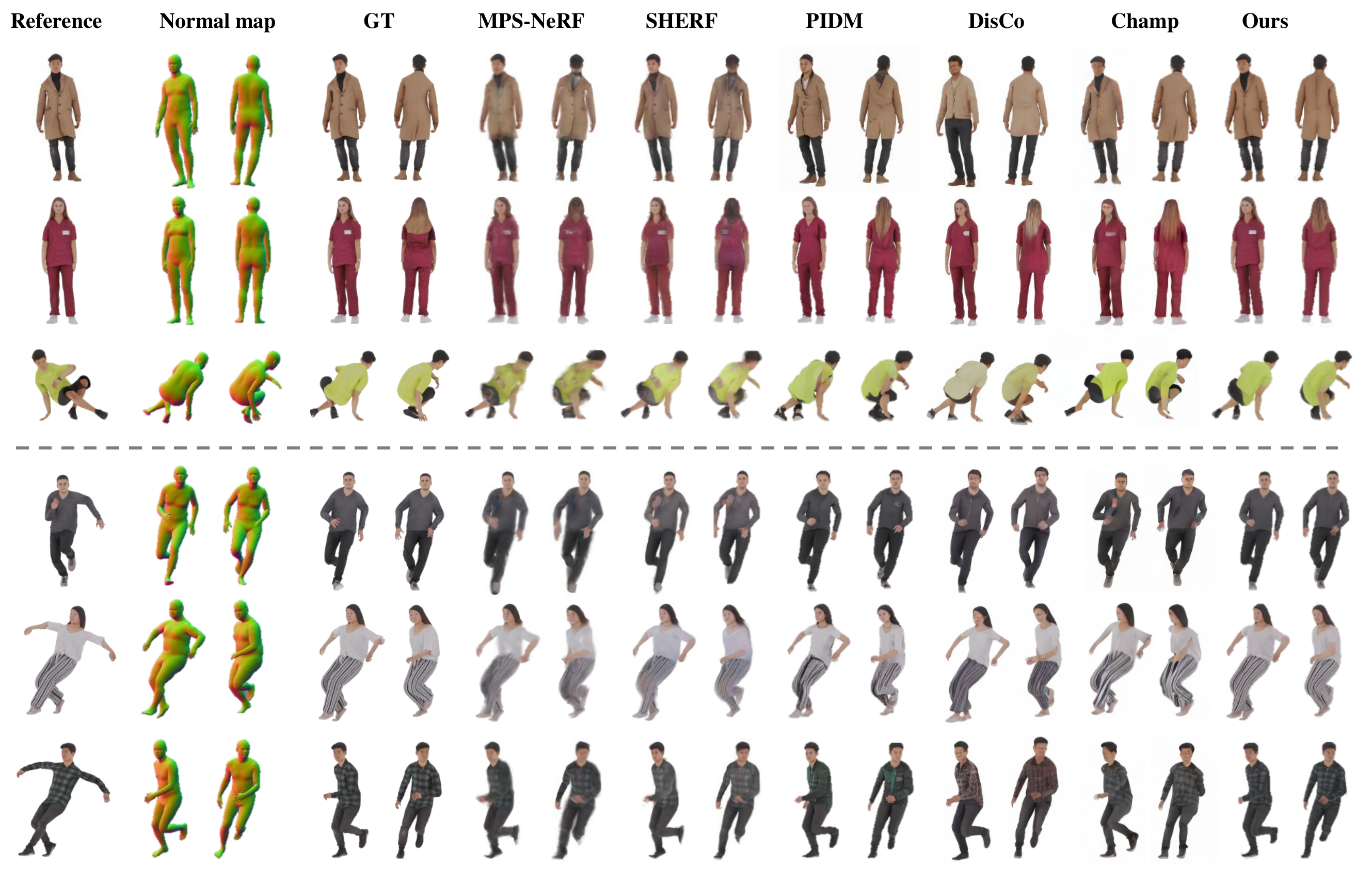}
				\caption{\red{Qualitative comparison to the state-of-the-art methods. We visualize the novel view  (row 1$\sim$ 3) and novel pose (row 4$\sim$ 6) results of MPS-NeRF \cite{MPS_NeRF} , SHERF \cite{SHERF}, PIDM \cite{PIDM}, Disco \cite{disco}, Champ \cite{champ} and our method on the RenderPeople dataset. Zoom in for better visualization. }
				}
				\label{fig:vision_results}
			\end{center}
		\end{figure}
		
		\myparagraph{Comparisons on RenderPeople \cite{renderpeople}.} From the results in Tab. \ref{table:STOA}, we have the following observations:
		(1) Generally, NeRF based methods (MPS-NeRF \cite{MPS_NeRF} and SHERF \cite{SHERF}) show higher PSNR and SSIM values than pose transfer based methods, as they are capable of preserving more accurate structures and textures for visible parts; while pose transfer based methods show lower LPIPS and FID values, as they are able to preserve high visual quality. \red{Additionally, when compared to other pose transfer-based methods that rely solely on ambiguous 2D pose guidance, such as DisCo, PoCoLD, and PIDM, Champ exhibits superior performance. This is attributed to its utilization of normal maps, semantic maps, and depth maps. However, it's challenging to ensure the accuracy of these maps simultaneously, which can potentially impact the generation results. Moreover, the modules introduced in Champ primarily emphasize the motion sequence within a video under a fixed viewpoint. Consequently, the consistent representation of characters from novel viewpoints might not be supported to preserve precise structures and textures under new viewpoints. 
		} In contrast, our method shows better numbers across all metrics. Specifically, our method reaches 23.29 w.r.t. PSNR, 0.91 w.r.t. SSIM and 0.07 w.r.t. LPIPS under the novel view scenario, outperforming previous methods by a large margin. These results indicate that our method generates images of both high accuracy and visual quality.
		(2) Image synthesis under novel pose is more challenging than that under novel view, but our method achieves a significant improvement of around 50\% (0.13 vs. 0.07) w.r.t. LPIPS compared to the previous best method SHERF, demonstrating that our method is able to generate high-quality images under the challenging novel pose scenario.

		Additionally, Fig. \ref{fig:vision_results} shows some qualitative comparison results on the RenderPeople dataset.
		We can see that NeRF based methods (MPS-NeRF \cite{MPS_NeRF} and SHERF \cite{SHERF}) make accurate predictions for those parts visible on the reference image, but fail at those invisible parts, resulting in blurry and even wrong details. For example, in row 1, both MPS-NeRF \cite{MPS_NeRF} and SHERF \cite{SHERF} wrongly predict the closure at the back of the coat; 
		similarly in row 2, they mistakenly put the tag at the back of the shirt. 
		These mistakes are caused by the missing correspondence between the target and reference images at those invisible parts, where the extracted features are inaccurate. 
		\red{The pose transfer based methods (PIDM, DisCo and Champ) output images of higher visual quality w.r.t. smoothness, but some details are inaccurate. For example, the face looks different from the reference for each person; }
		\red{Also, PIDM and DisCo usually fail under complex poses with severe occlusions, due to the lack of 3D information. } For example, 
		in row 3, 
		the right arm is broken, being mixed with the right leg. \red{Although Champ benefits from the assistance of normal maps, semantic maps, and depth maps, ensuring the accuracy and consistency of these maps simultaneously remains a challenging task, ultimately leading to incorrect hand posture in row 4 and foot posture in row 6.}
		In contrast, our results are of high visual quality and provide more accurate details (e.g. face, hair) even at those invisible body parts.
		
		\setlength{\tabcolsep}{2pt}
		\begin{table}[t]
			
			\renewcommand*{\arraystretch}{1.15}
			\centering
			\begin{tabular}{l|cccc|cccc}
				\toprule
				
				\multirow{2}{*}{Method} &\multicolumn{4}{c|}{Novel View}&\multicolumn{4}{c}{Novel Pose}\\
				\cline{2-9} 
				~ &FID$\downarrow$ &PSNR$\uparrow$ &SSIM$\uparrow$& LPIPS$\downarrow$ &FID$\downarrow$&PSNR$\uparrow$ &SSIM$\uparrow$& LPIPS$\downarrow$ \\
				\hline
				MPS-NeRF&60.74&18.20 &0.82& 0.16 & 62.72 &17.61& 0.81 & 0.17\\
				SHERF & 37.76& \textbf{20.37}& \textbf{0.88} & 0.11 & 37.48 & 18.85& 0.85 & 0.12 \\ 
				\red{DisCo} &\red{28.71}& \red{15.03} & \red{0.72} & \red{0.14}&\red{34.63}& \red{13.08} & \red{0.67} & \red{0.18} \\
				\red{Champ}&\red{36.75}& \red{16.24}& \red{0.75}& \red{0.13}&\red{41.23}& \red{15.96} &\red{0.74} &\red{ 0.14} \\
				PoCoLD&24.81 & 16.53 & 0.79 & 0.13&27.87 & 15.88 & 0.77 &0.14 \\ 
				PIDM&32.45 & 16.23 & 0.79 & 0.14&35.05 & 16.09 & 0.78 &0.14 \\ 
				\hline
				Ours & \textbf{18.22} &19.89 & 0.87 & \textbf{0.08} & \textbf{21.82}& \textbf{19.60} & \textbf{0.87} &\textbf{0.08}\\ 
				\toprule
			\end{tabular}
			\caption{\red{Quantitative comparison to the state-of-the-arts on the THuman dataset. \fixx{$\uparrow$: higher is better. $\downarrow$: lower is better.}}}
			\label{table:STOA_thuman}
		\end{table}
		
		\begin{figure}[t]
			\begin{center}
				\includegraphics[width=0.99\textwidth]{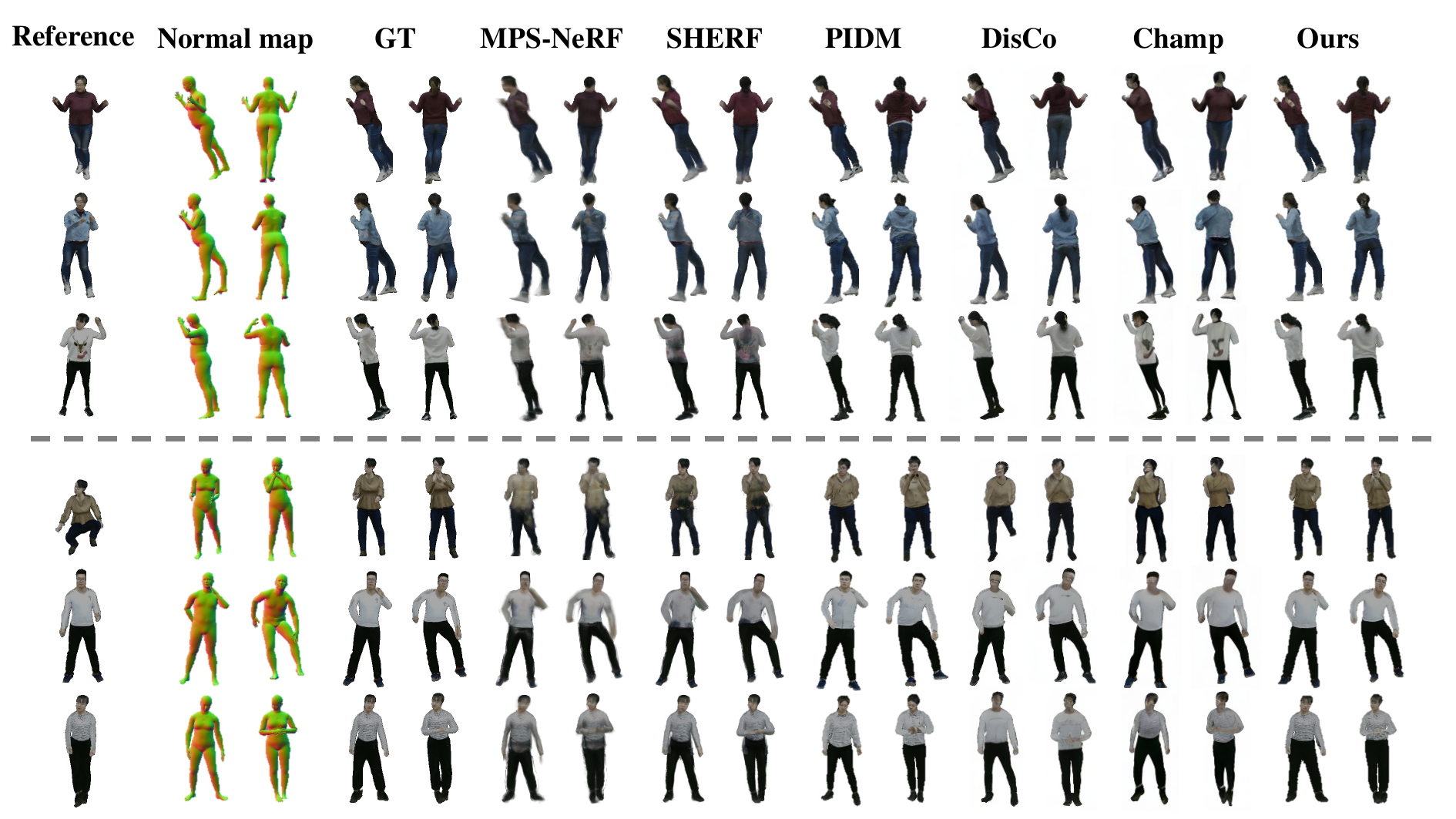}
				\caption{\red{Qualitative comparison to the state-of-the-art methods. We visualize the novel view  (row 1$\sim$ 3) and novel pose (row 4$\sim$ 6) results of MPS-NeRF \cite{MPS_NeRF} , SHERF \cite{SHERF}, PIDM \cite{PIDM}, Disco \cite{disco}, Champ \cite{champ} and our method on the THuman \cite{thuman} dataset. Zoom in for better visualization. }
				}

				\label{fig:vision_sup_thuman}
			\end{center}
		\end{figure}

		\myparagraph{Comparisons on THuman \cite{thuman}.} We list the quantitative comparison results on THuman in Tab. \ref{table:STOA_thuman}. Since we render the training and test data from their coarse 3D scans and the quality of rendered images is low, THuman data is more challenging than RenderPeople data.
		In the task of novel-view image synthesis, both FID and LPIPS of our method are much lower than that of the compared methods which indicates our method can generate higher visual quality images. Only the PSNR and SSIM accuracy of SHERF \cite{SHERF} is a bit higher than our method because SHERF may recover more accurate pixel-to-pixel correspondences with the ground truth image while our method mainly focuses on enhancing image-level visual quality and may ignore recovering precise pixel-level correspondences. In the more difficult task of novel-pose image synthesis, our method achieves state-of-the-art performance in both synthesis accuracy and visual quality. \red{Fig. \ref{fig:vision_sup_thuman} shows some qualitative comparison results on the
			THuman dataset, showing our method generates human images of both higher accuracy and better visual quality in varying views and poses. }



		\subsection{Ablation Studies}
		In the following, we conduct ablation studies to demonstrate the effectiveness of different components of our method. All experiments are conducted on the RenderPeople dataset.
		
		\setlength{\tabcolsep}{5pt}

		\begin{table*}[t]
			
			\renewcommand*{\arraystretch}{1.05}
			\centering
			\begin{tabular}{ccc|ccc|ccc}
				\toprule
				\multirow{2}{*}{3DN} &\multirow{2}{*}{CP}&\multirow{2}{*}{SR}&\multicolumn{3}{c|}{Novel View} &\multicolumn{3}{c}{Novel Pose} \\ 
				\cline{4-9}
				~&~ & ~ &PSNR$\uparrow$ &SSIM$\uparrow$& LPIPS$\downarrow$ &PSNR$\uparrow$ &SSIM$\uparrow$& LPIPS$\downarrow$ \\
				\hline
				$\times$ & $\times$& $\times$ & 19.82 & 0.85 & 0.11 & 19.12 & 0.83 &0.12\\  
				\checkmark & $\times$& $\times$ & 21.50 & 0.88 & 0.10 & 20.83 & 0.87 &0.11\\  
				$\times$ & \checkmark & $\times$&  \fix{20.98} &\fix{ 0.87} & \fix{0.10} & \fix{20.34} & \fix{0.86} &\fix{0.10}\\  
				\checkmark & \checkmark& $\times$ & 21.61 & 0.89 & 0.09 & 20.96 & 0.87 &0.10\\  
				$\times$& $\times$ & \checkmark & 21.72 & 0.88 & 0.09 & 20.59 & 0.86 &0.11\\  
				$\times$ & \checkmark & \checkmark & \fix{22.74} & \fix{0.90} & \fix{0.07} & \fix{21.81} & \fix{0.88} &\fix{0.08}\\  
				\checkmark & \checkmark & \checkmark& 23.29 & 0.91 & 0.07 & 22.20 & 0.90&0.08\\     
				\toprule
			\end{tabular}
			\caption{\textcolor{black}{Impact of 3D prior guidance (3D normal map (3DN) and color prompt (CP)) and self-reconstruction based customized refinement (SR).}}
			\label{table:ablation}
		\end{table*}
		
		\myparagraph{Impact of 3D prior guidance and self-reconstruction based customized refinement.}
		As shown in Tab. \ref{table:ablation}, \fix{the 3D normal map alone improves PSNR by 1.68 (from 19.82 to 21.50), while the color prompt independently contributes 1.16 PSNR gain (from 19.82 to 20.98), indicating larger accuracy improvement of using the 3D normal map over the color prompt.} The performance is improved by 1.79 pp w.r.t. PSNR (from 19.82 to 21.61) when 3D human prior guidance is introduced to novel view synthesis. On the other hand, we also observe a remarkable improvement of 1.90 pp w.r.t. PSNR (from 19.82 to 21.72) from self-reconstruction based customized refinement. Moreover, we obtain a total gain of 3.47 pp w.r.t. PSNR by activating both of them.  Our approach can boost improvements in both novel view and novel pose generation, indicating the effectiveness of two proposed technics.  \fix{And removing the 3D normal map from our full model reduces PSNR by 0.55 (from 23.29 to 22.74), which demonstrates introducing the 3D normal map can enhance the synthesis quality obviously and confirms its critical role in geometry-aware synthesis.}
		
\begin{figure}[t]
			\begin{center}
				\includegraphics[width=0.95\textwidth]{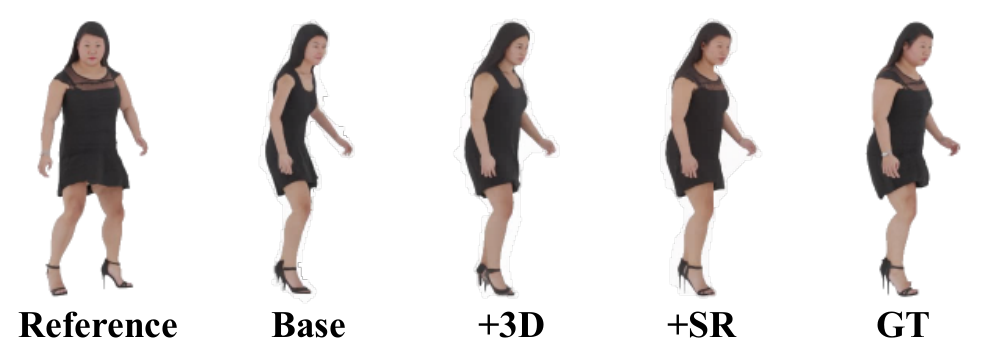}
			\end{center}
			\caption{\textcolor{black}{Step-by-step improvement of our method, indicating the effects of both 3D prior guidance and self-reconstruction based customized refinement}.} 
			\label{fig:ablation}
		\end{figure}

		Additionally, one visualization example is shown in Fig. \ref{fig:ablation}. Adding the 3D guidance, the body shape is more consistent with the reference image. Combining both 3D human prior and self-reconstruction can further improve the generated images with fine-grained details.
		
		\setlength{\tabcolsep}{2pt}
		\begin{table}[tb]
			
			\renewcommand*{\arraystretch}{1.15}
			\centering
			\begin{tabular}{l|ccc|ccc}
				\toprule
				\multirow{2}{*}{Method}&\multicolumn{3}{c|}{Novel View}&\multicolumn{3}{c}{Novel Pose}\\
				\cline{2-7} 
				~ &PSNR$\uparrow$ &SSIM$\uparrow$& LPIPS$\downarrow$ &PSNR$\uparrow$ &SSIM$\uparrow$& LPIPS$\downarrow$ \\
				\hline
				SHERF\cite{SHERF} & 21.37 & 0.88 & 0.12 & 20.65 & 0.86 & 0.13 \\ 
				SHERF(w/ SR) & 21.92 & 0.88 & 0.12 & 21.10 & 0.87 & 0.13 \\ 
				\hline
				PIDM\cite{PIDM} & 19.82 & 0.85 & 0.11 & 19.12 & 0.83 & 0.12 \\ 
				PIDM(w/ SR) & 21.72 & 0.88 & 0.09 & 20.59 & 0.86 &0.11\\ 
				\hline
				Ours(w/o SR) & 21.61 & 0.89 & 0.09 & 20.96 & 0.87 &0.10\\
				Ours(w/ SR) & \textbf{23.29} & \textbf{0.91} & \textbf{0.07} & \textbf{22.36} & \textbf{0.90} &\textbf{0.07}\\ 
				\toprule
			\end{tabular}
			\caption{\textcolor{black}{Our proposed self-reconstruction based customized refinement (SR) is generalizable across different generators.}}
			\label{table:self-generation}
		\end{table}

		\myparagraph{Generalization ability of self-reconstruction based customized refinement.}
		In the inference phase, our self-reconstruction exploits the reference information for detail preservation. It can be considered as a plug-in step and combine it with prior methods. From the results reported in Tab. \ref{table:self-generation}, \fix{without SR, there is a minimal difference (0.24 PSNR gap) between our method and SHERF. Activating SR leads to a 1.9 PSNR gain for PIDM and 1.68 PSNR for our method, while SHERF only improves by 0.55 PSNR. Therefore, the integration of SR yields an obvious improvement for these methods and the comparison results demonstrates the stronger compatibility of SR with our model. }
		We have the following observations.
		(1) Different methods all benefit from our proposed self-reconstruction based customized refinement, demonstrating its generalization ability. 
		(2) The improvement over the NeRF-based method is rather limited. The reason is that the pixel-aligned features used in NeRF inference are able to preserve visible details to a certain extent. 
		(3) Our method unleashes its full potential and outperforms other methods once self-reconstruction is introduced.
		Since our diffusion based model is of high capacity and easy to overfit on the small training dataset, the self-reconstruction based customized refinement successfully overcomes this problem and brings significant improvement.
		
		\setlength{\tabcolsep}{2pt}
		\begin{table}[b]
			
			
			\renewcommand*{\arraystretch}{1.15}
			\centering
			\begin{tabular}{l|ccc|ccc}
				\toprule
				\multirow{2}{*}{Method}&\multicolumn{3}{c|}{Novel View}&\multicolumn{3}{c}{Novel Pose}\\
				\cline{2-7} 
				~ &PSNR$\uparrow$ &SSIM$\uparrow$& LPIPS$\downarrow$ &PSNR$\uparrow$ &SSIM$\uparrow$& LPIPS$\downarrow$ \\
				\hline
				MPS-NeRF & 15.59 & 0.77 & 0.20 & 15.42 & 0.76 & 0.20 \\ 
				SHERF & 17.98 &0.81 & 0.15 & 17.28 & 0.80 & 0.16 \\ 
				PIDM & 14.83& 0.73&0.18  & 14.61 & 0.72 & 0.19\\ 
				\hline
				Ours & \textbf{18.47} & \textbf{0.84} & \textbf{0.10} & \textbf{17.94} &\textbf{0.83}&\textbf{0.11} \\
				\toprule
			\end{tabular}
			\caption{\textcolor{black}{Quantitative results for cross-dataset (RenderPeople $\rightarrow$ THuman) validation.}
			}\label{table:cross-dataset}
			
		\end{table}
		
\begin{figure}[t]
			\begin{center}
				\includegraphics[width=1.0\textwidth]{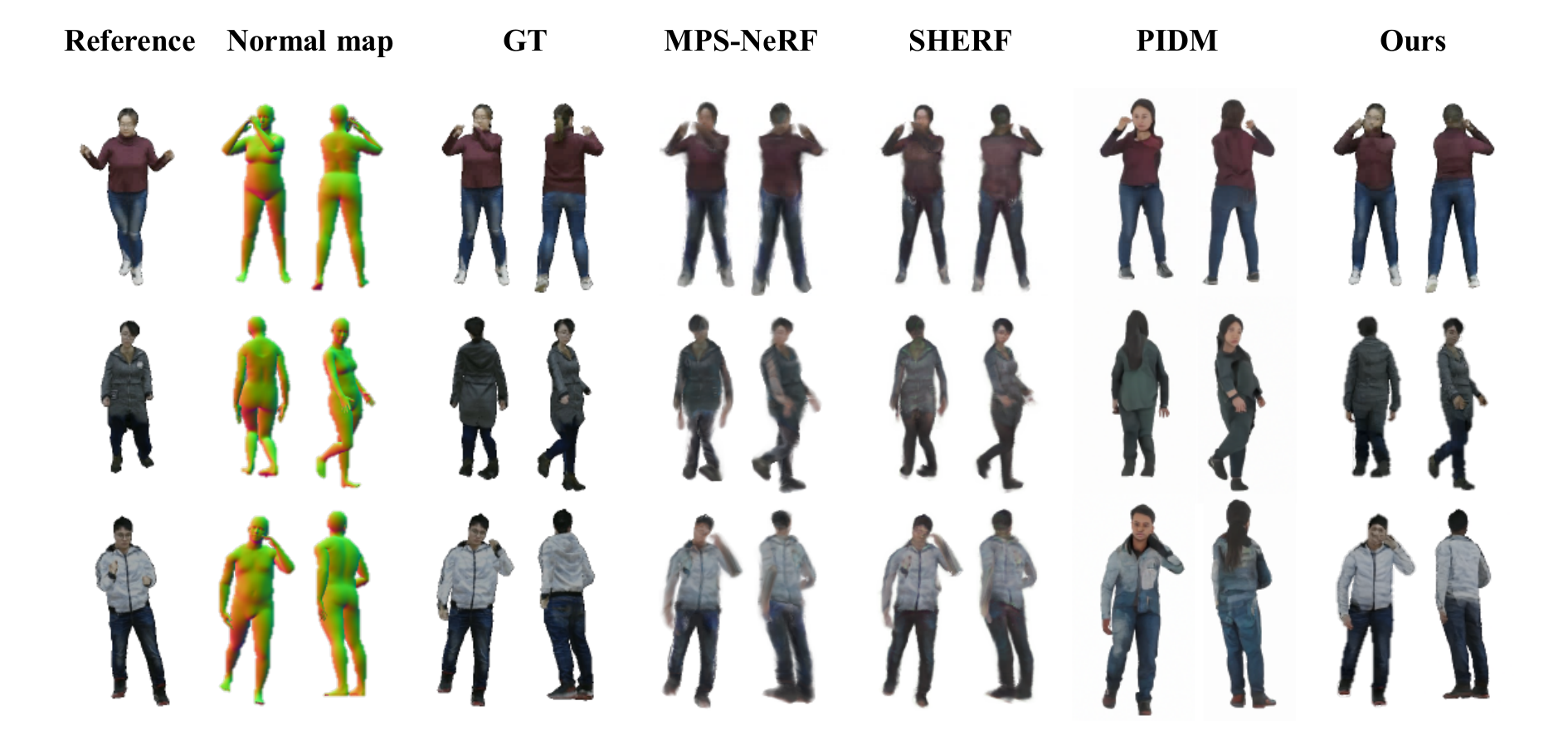}
				\caption{\textcolor{black}{Qualitative results for cross-dataset (RenderPeople $\rightarrow$ THuman) validation. Better visual results indicate better cross-dataset generalization ability. Zoom in for better visualization.}
				}
				\label{fig:cross-dataset}
			\end{center}
		\end{figure}

		\myparagraph{Analysis on generalization ability.} 
		To further study the generalization ability of our method, we perform cross-dataset validation by applying RenderPeople-trained models of different methods on the THuman dataset. As shown in Tab. \ref{table:cross-dataset}, our method outperforms other approaches across all three metrics, especially with notable improvements w.r.t. LPIPS. These results indicate that our method generalizes better to a novel dataset, and thus is more robust to variations in terms of camera, illumination and so on. We also show some qualitative results in Fig. \ref{fig:cross-dataset}, where we can see our method generates images of higher visual quality and fidelity than previous methods.
		
		\setlength{\tabcolsep}{5pt}
		\begin{table}[t]

			\renewcommand*{\arraystretch}{1.15}
			\centering
			\begin{tabular}{c|ccc|ccc}
				\toprule
				\multirow{2}{*}{$\bm{\lambda}$}&\multicolumn{3}{c|}{Novel View}&\multicolumn{3}{c}{Novel Pose}\\
				\cline{2-7} 
				~ &PSNR$\uparrow$ &SSIM$\uparrow$& LPIPS$\downarrow$ &PSNR$\uparrow$ &SSIM$\uparrow$& LPIPS$\downarrow$ \\
				\hline
				0 & 21.61 & 0.89 & 0.09 & 20.96 & 0.87 &0.10\\
				50 & 23.23 & 0.91 & 0.07 & 22.24 & 0.89 &0.08\\
				100 & \textbf{23.29} & \textbf{0.91} & \textbf{0.07} & \textbf{22.36} & \textbf{0.90} &\textbf{0.07}\\ 
				200 & 23.20 & 0.91 & 0.07 & 22.25 & 0.90 &0.08\\ 
				\toprule
			\end{tabular}
			\caption{The effects of different re-optimization steps $\lambda$ on the synthesis performance evaluated on the RenderPeople dataset\cite{renderpeople}.}\label{table:optimization step}
		\end{table}
		
		\setlength{\tabcolsep}{2pt}
		\begin{table}[tb]
			
			\renewcommand*{\arraystretch}{1.15}
			\centering
			\begin{tabular}{l|c|ccc|ccc}
				\toprule
				\multirow{2}{*}{Method}&\multirow{2}{*}{\shortstack{GT \\ Condition?}}&\multicolumn{3}{c|}{Novel View}&\multicolumn{3}{c}{Novel Pose}\\
				\cline{3-8} 
				~ &~&PSNR$\uparrow$ &SSIM$\uparrow$& LPIPS$\downarrow$ &PSNR$\uparrow$ &SSIM$\uparrow$& LPIPS$\downarrow$ \\
				\hline
				\multirow{2}{*}{MPS-NeRF\cite{MPS_NeRF}} &\checkmark& 20.36 & 0.84 & 0.16 & 19.61 & 0.82 & 0.18 \\ 
				~ & $\times$& 17.96 & 0.76 & 0.21 & 16.24 & 0.72 & 0.25 \\ 
				\hline
				\multirow{2}{*}{SHERF\cite{SHERF}} &\checkmark& 21.37 & 0.88 & 0.12 & 20.65 & 0.86 & 0.13 \\ 
				~ & $\times$& 17.70 & 0.77 & 0.18 & 15.89 & 0.71 & 0.23 \\ 
				\hline
				\multirow{2}{*}{PIDM\cite{PIDM}}&\checkmark & 19.82 & 0.85 & 0.11 & 19.12 & 0.83 & 0.12 \\ 
				~& $\times$ & 17.73 & 0.77 & 0.15 & 17.08 & 0.74 &0.16\\ 
				\hline
				\multirow{2}{*}{Ours}&\checkmark& 23.29 & 0.91 & 0.07 & 22.36 & 0.90 &0.07\\
				~ & $\times$& 18.59 & 0.80 & 0.12 & 18.09 & 0.78 &0.13\\ 
				\toprule
			\end{tabular}
			\caption{\fix{The effects of guidance condition accuracy on the RenderPeople dataset\cite{renderpeople}. \checkmark denotes the use of ground-truth (GT) conditions, while the $\times$ indicates the utilization of predicted conditions derived from the HMR2.0 \cite{HMR2}.}}
			\label{table:HMR}
		\end{table}

		\myparagraph{Impact of re-optimization steps $\lambda$.} We analyze how the re-optimization step $\lambda$ in our self-reconstruction method affects the synthesis performance by evaluating with different values of $\lambda$ on the RenderPeople dataset\cite{renderpeople}. As shown in Tab. \ref{table:optimization step}, the synthesis performance is improved after employing self-reconstruction based customized refinement. Since our method obtains the best performance using $\lambda=100$, we choose $\lambda=100$ for all our experiments. \fix{It requires approximately 90s on RTX 4090 Ti. Crucially, this optimization is only need to be performed once per target identity, and the efficient subsequent generation is enabled without per-image re-optimization based on the refined model.}

		
		\myparagraph{\fixx{GT vs. Estimated SMPL parameters}} \fixx{To analyze the impact of guidance condition accuracy, we replace the GT SMPL parameters with HMR2.0 \cite{HMR2} predictions during image generation. As demonstrated in Tab. \ref{table:HMR}, when ground-truth SMPL parameters are replaced with estimated parameters from HMR2.0, the performance of all methods degrades noticeably, and our method also observes a clear drop in quantitative metrics. The clear drop for all methods may be due to the complex human poses and serious occlusions in the test samples, posing challenges on the SMPL estimation. The prediction errors of human body shapes and poses lead to inaccurate guidance conditions, and further affect the quality of image synthesis. Although the synthesis quality and fidelity of our method are lower than that using GT SMPL guidance, our method still outperforms other methods under estimated SMPL conditions. Overall, the in-the-wild application of our method is bounded by the accuracy of off-the-shelf SMPL estimators, and further improvements in real-world scenarios can be achieved by adopting a more robust and accurate SMPL estimator.}

\begin{figure}[H]
			\begin{center}
				\includegraphics[width=0.95\textwidth]{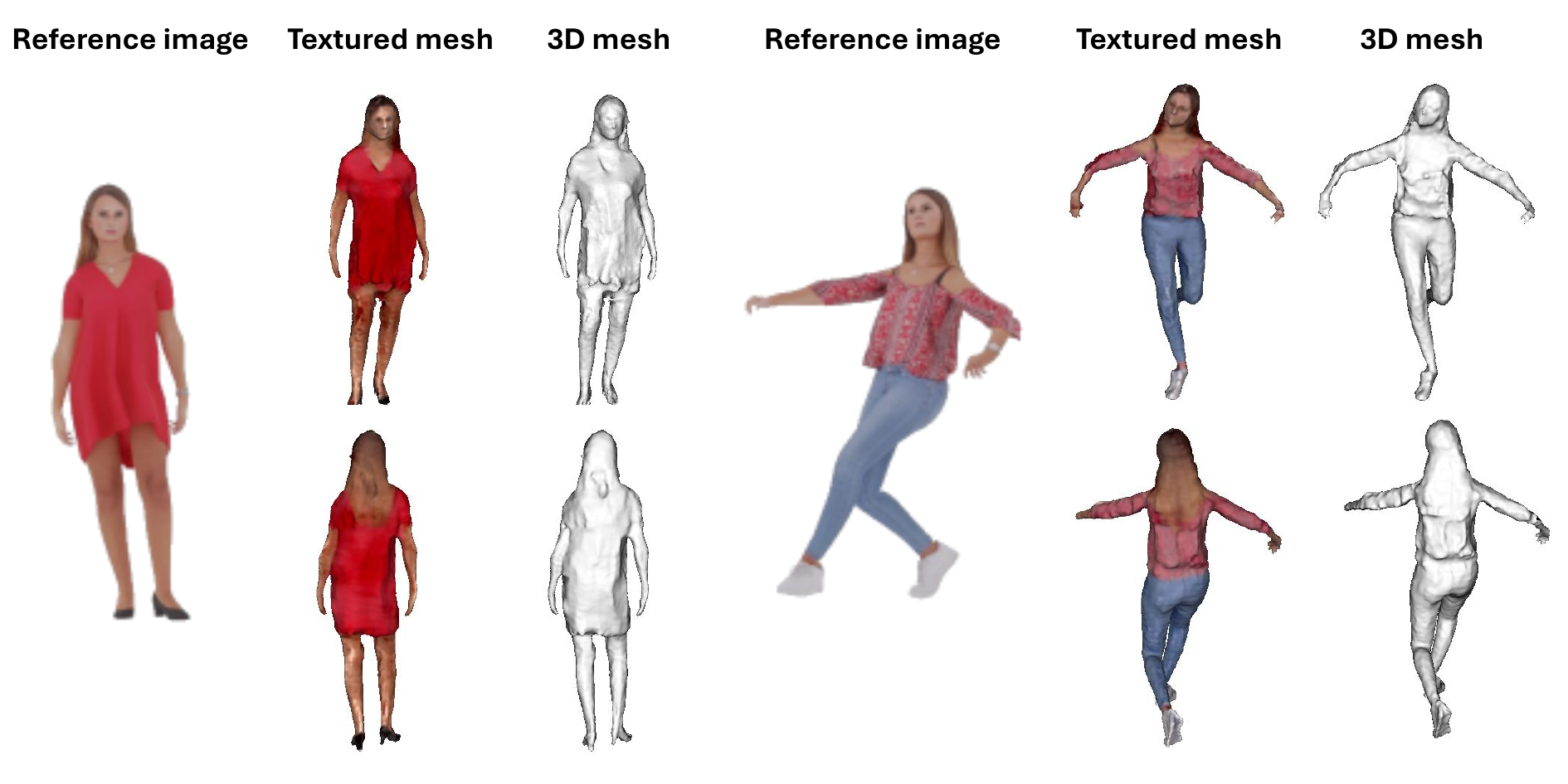}
			\end{center}
			\caption{\textcolor{black}{Application of single-image 3D human reconstruction using our method on RenderPeople \cite{renderpeople}. For each person, we show the input reference image and reconstruction results in two different views. The 3D human meshes are recovered with about 15 images from different views generated using our method.}} 
			\label{fig:3drecon}
		\end{figure}

		\myparagraph{Application: Single-image 3D human reconstruction.}
		Benefiting from the generated multi-view images using our method, we can enable interesting applications like single-image 3D human reconstruction as shown in Fig. \ref{fig:3drecon}. From a reference human image, we first generate multi-view human images from different views, and then reconstruct the detailed 3D humans using the multi-view reconstruction method \cite{3drec}. In the examples of Fig. \ref{fig:3drecon} reconstructed using about 15 different views, we can see the human details in 3D meshes are recovered successfully, which demonstrates that the synthesized multi-view images are consistent across different views with high visual quality.

		\section{Conclusion}
		In this work, we aim to synthesize novel view and pose human images of both high visual quality and fidelity, given only one single reference image.
		We first build a new generator by introducing 3D human priors as additional conditions to a denoising diffusion model, and then update the generator by a self-reconstruction based customized refinement when applying it to a novel unseen person so that more high-frequency details can be well recovered.
		Experimental results show that our method outperforms previous methods w.r.t. both visual quality and fidelity, and is also more generalizable to a novel dataset.
		
		\section*{Acknowledgment}
		The authors would like to thank the editor and the anonymous reviewers for their critical and constructive comments and suggestions. This work was supported in part by the National Natural Science Foundation of China (Grant No. 62172225, No. 62322602, No. 62472224 \& No. 62506373), the Natural Science Foundation of Jiangsu Province, China (Grant No. BK20230033) and the China Postdoctoral Science Foundation (Grant No. 2025M784475).
		
		%
		%
		
		
		
		\bibliographystyle{IEEEtran}
		\bibliography{egbib}
		
	\end{document}